\documentclass[10pt,twocolumn,letterpaper]{article}

\usepackage[pagenumbers]{cvpr} 

\usepackage{graphicx}
\usepackage{amsmath}
\usepackage{amssymb}
\usepackage{booktabs}
\usepackage{amsfonts,amssymb}
\usepackage{mathrsfs}

\usepackage{mathrsfs}
\usepackage[pagebackref,breaklinks,colorlinks]{hyperref}
\usepackage{multirow}
\usepackage{graphicx}
\usepackage[table,xcdraw]{xcolor}
\usepackage{bm}

\usepackage[capitalize]{cleveref}
\crefname{section}{Sec.}{Secs.}
\Crefname{section}{Section}{Sections}
\Crefname{table}{Table}{Tables}
\crefname{table}{Tab.}{Tabs.}

\def\cvprPaperID{3405} 
\def\confName{CVPR}
\def\confYear{2022}

\begin{document}

\title{Pyramid Grafting Network for One-Stage High Resolution \\
Saliency Detection}

\author{Chenxi Xie\textsuperscript{1}, Changqun Xia\textsuperscript{*2}, Mingcan Ma\textsuperscript{1}, Zhirui Zhao\textsuperscript{1}, Xiaowu Chen\textsuperscript{1,2}, Jia Li\textsuperscript{1,2} \\
\textsuperscript{1}State Key Laboratory of Virtual Reality Technology and Systems, SCSE, Beihang University\\
\textsuperscript{2}Peng Cheng Laboratory, Shenzhen, China\\
{\tt\small \{xiechenxi,mingcanma,zhiruizhao,chen,jiali\}@buaa.edu.cn, xiachq@pcl.ac.cn}}


\maketitle

\begin{abstract}
   Recent salient object detection (SOD) methods based on deep neural network have achieved remarkable performance. However, most of existing SOD models designed for low-resolution input perform poorly on high-resolution images due to the contradiction between the sampling depth and the receptive field size. Aiming at resolving this contradiction, we propose a novel one-stage framework called Pyramid Grafting Network (PGNet), using transformer and CNN backbone to extract features from different resolution images independently and then graft the features from transformer branch to CNN branch. An attention-based Cross-Model Grafting Module (CMGM) is proposed to enable CNN branch to combine broken detailed information more holistically, guided by different source feature during decoding process. Moreover, we design an Attention Guided Loss (AGL) to explicitly supervise the attention matrix generated by CMGM to help the network better interact with the attention from different models. We contribute a new Ultra-High-Resolution Saliency Detection dataset UHRSD, containing 5,920 images at 4K-8K resolutions. To our knowledge, it is the largest dataset in both quantity and resolution for high-resolution SOD task, which can be used for training and testing in future research. Sufficient experiments on UHRSD and widely-used SOD datasets demonstrate that our method achieves superior performance compared to the state-of-the-art methods.

\end{abstract}

\renewcommand{\thefootnote}{}
\footnote{ \textsuperscript{*}Correspondence should be addressed to Changqun Xia (Email: \href{mailto:xiachq@pcl.ac.cn}{xiachq@pcl.ac.cn} ). The code and dataset are available at \url{https://github.com/iCVTEAM/PGNet}.  }
\section{Introduction}
Salient object detection (SOD) \cite{cheng2014global,borji2019salient} aims at identifying and segmenting the most attractive objects in a certain scene. As a pre-processing step, it is widely applied in various computer vision tasks, such as light field segmentation\cite{liu2021light,zhang2019memory}, instance segmentation \cite{zhou2020multi} and video object segmentation\cite{ji2021full,zhang2021dynamic}.

\begin{figure}[t]
\centering
\includegraphics[width=1.0\columnwidth]{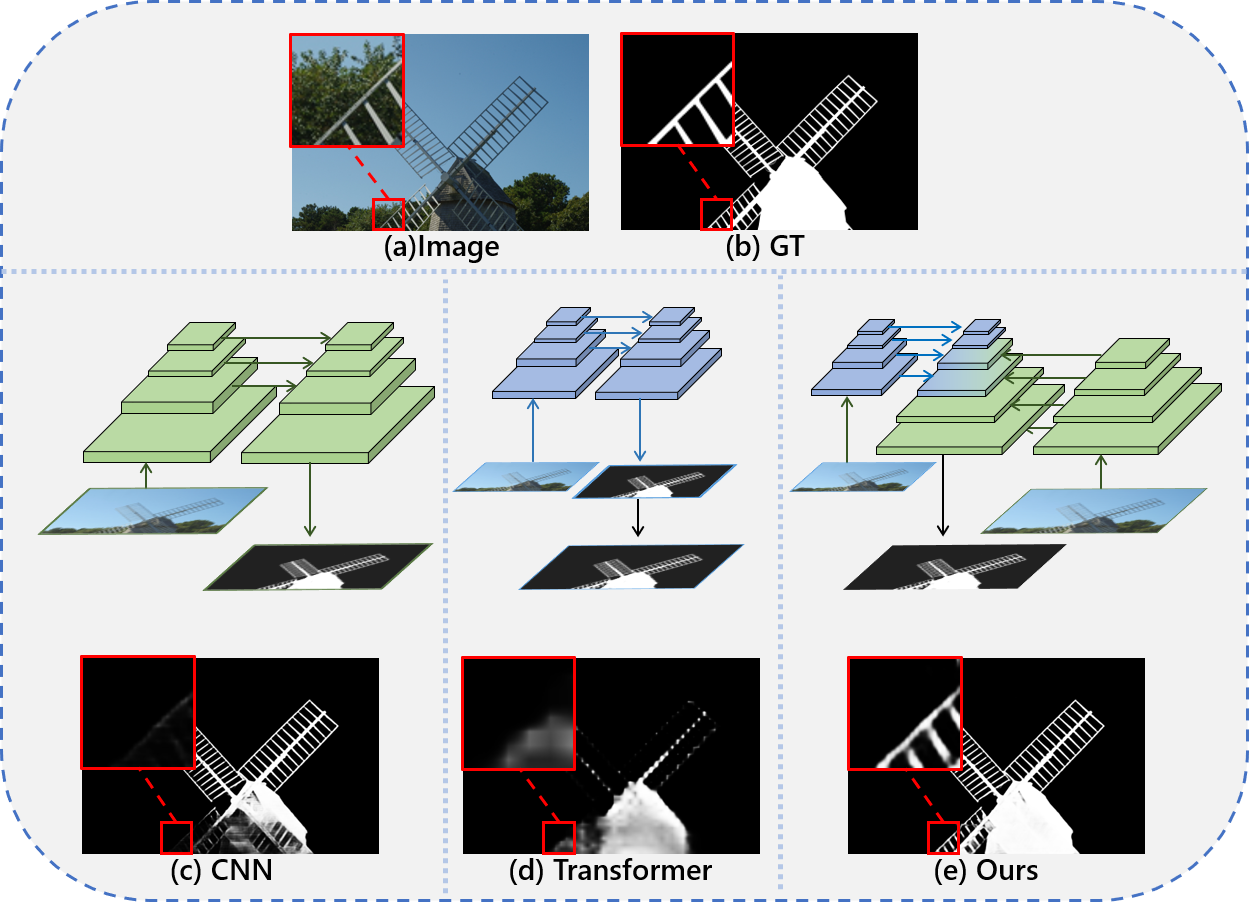}
\caption{Comparison of the results of the different methods. (a) Input image. (b) Ground truth mask. (c) Directly input to Resnet-18 based FPN. (d) Downsample then input to Swin transformer based FPN. (e) Ours. }
\label{fig:first}
\end{figure}

Recently, deep neural networks based salient object detection methods have made remarkable achievements\cite{chen2018reverse,su2019banet,liu2020dynamic,qin2020u2,ji2021calibrated,fan2020bbs}. However, most of existing SOD methods perform well within a specific input low-resolution range (\eg, $224 \times 224, 384\times 384$ ). With the rapid development of image capture devices (\eg., smartphone), the resolution (\eg, 1080p, 2K and 4K) of the images accessible to people has far exceeded the range to which existing saliency detection method can be adapted directly. As shown in \cref{fig:first} (c), we fed the high-resolution image directly into the commonly used network with Resnet-18 as the backbone, and comparing ground truth \cref{fig:first} (b) shows that the segmentation result is incomplete and many detail regions are lost. In order to reduce computational consumption and memory usage, existing methods often downsample the input images and then upsample the output results to recover original resolution, as illustrated in \cref{fig:first} (d). This challenge is due to the fact that most low-resolution SOD networks are designed in an Encoder-Decoder style, and as the input resolution increases dramatically, the size of features extracted increases, but the receptive field determined by the network is fixed, making the relative receptive field small, ultimately resulting in the inability to capture global semantics that are vital to SOD task. 
Since direct processing cannot handle the challenges posed by high resolution, a number of methods have emerged in recent years specifically designed for high-resolution input.
There are two representative high-resolution SOD methods (HRSOD\cite{zeng2019towards}, DHQSOD\cite{tang2021disentangled}). HRSOD divides the whole process into global stage, local stage and reorganization stage, where the global stage provides guidance on both the local stage and the crop process. And DHQSOD disentangle the SOD task into classification task and regression task, where the two task is connected by their proposed trimap and uncertainty loss. They generate relatively good saliency maps with sharp boundaries. 

However, both of the above methods use a multi-stage architecture, dividing SOD into semantic(in low resolution) and detailed (in high resolution) phases. This has led to two new problems: (1) Inconsistent contextual semantic transfer between stages. The intermediate maps obtained in the previous stages are input into the last stage, while the errors are also passed on. Further more, the refinement in the last stage will likely inherit or even amplify the previous errors as there is not enough semantic support, which implies that the final saliency maps are heavily dependent on the performance of the low-resolution network. (2) Time consuming. Compared to the one-stage method, the multi-stage method are not only difficult to parallel but also have the potential problem of increasing number of parameters, which makes it slow.

Based on the above defects of existing high-resolution methods , we propose a new perspective that since the specific features in a single network cannot settle the paradox of receptive field and detail retention simultaneously, instead we can separately extract two sets of features of different spatial sizes and then graft the information from one branch to the other. In this paper, we rethink the dual-branch architecture and design a novel one-stage deep neural network for high-resolution saliency detection named Pyramid Grafting Network (PGNet). As illustrated in \cref{fig:first} (e), we use both Resnet and Transformer as our Encoders, extracting features with dual spatial sizes in parallel. The transformer branch first decode the features in the FPN style, then pass the global semantic information to the Resnet branch in the stage where the feature maps of two branches have similar spatial sizes. We call this process feature grafting. Eventually, the Resnet branch completes the decoding process with the grafted features. Compared to classic FPNs, we have constructed a higher feature pyramid at a lower cost. To better graft features cross two different types of models, we design the Cross-Model Grafting Module (CMGM) based on the attention mechanism and propose the Attention Guided Loss to 
further guide the grafting. Considering that supervised deep learning method requires a large amount of high quality data, we have provided a 4K resolution SOD dataset (UHRSD) with the largest number to date in an effort to promote future high-resolution salient object detection research.

Our major contributions can be summarized as follows:

\begin{itemize}

\item We propose the first one-stage framework named PGNet for high-resolution salient object detection, which uses staggered connection to capture both continuous semantics and rich details.

\item We introduce the Cross-Model Grafting Module to transfer the information from transformer branch to CNN branch, which allows CNN to not only inherit global information but also remedy the defects common to both. Moreover, we design the Attention Guided Loss to further promote the feature grafting.

\item We contribute a new challenging Ultra High-Resolution Saliency Detection dataset (UHRSD) containing 5,920 images of various scenes at over 4K resolution and corresponding pixel-wise salient annotations, which is the largest high-resolution saliency dataset available.

\item Experimental results on both existing datasets and ours demonstrate our method outperforms state-of-the-art methods in accuracy and speed.

\end{itemize}

\label{sec:intro}

\section{Related work}
During the past decades, a large amount traditional methods have been proposed to  solve saliency detection problem\cite{yan2013hierarchical,itti1998model,yang2013saliency}.  However, these methods only focus on the low-level feature and ignore the rich semantic information resulting in unstable performance in complex scenarios.More details can be found in \cite{borji2019salient}.
\subsection{Deep Learning-Based Saliency Detection}
Recently, remarkable progress has been made in saliency detection due to the application of deep neural network \cite{ zhao2019egnet, wei2020label, lin2017feature,xia2017and, xu2021locate}. Hou \etal \cite{hou2017deeply} and Chen \etal \cite{chen2020global} use deep convolutional networks as Encoder to extract multi-level features and design various modules to fuse them in an FPN style. Ma \etal \cite{ma2021pyramidal} and Xu \etal \cite{xu2021locate} avoid semantic dilution while suppressing loss of detail by experimenting with various feature connection paths. In addition, Wei \etal \cite{wei2020label} generate saliency maps with sharp boundary by explicitly supervising edge pixels. The extensive use of transfomer in vision has also led to new advances in saliency detection. Liu \etal \cite{liu2021visual} take use of T2T-vit as 
backbone and design a multi-tasking decoder with a pure transformer architecture to perform RGB and RGB-D saliency detection. However, these methods are designed for low-resolution scenes and cannot be directly applied to high-resolution scenes.

\subsection{High-Resolution SOD}
Nowadays, focusing on high-resolution SOD methods is already trending.
Zeng \etal \cite{zeng2019towards} propose a paradigm for high-resolution salient object detection using GSN for extracting semantic information, and APS guided LRN for optimizing local details and finally GLFN for prediction fusion. Also they contributed the first high-resolution salient object detection dataset (HRSOD). Tang \etal \cite{tang2021disentangled} propose that salient object detection should be disentangled into two tasks. They first design LRSCN to capture sufficient semantics at low resolution and generate the trimap. By introducing the uncertainty loss, the designed HRRN can refine the trimap generated in first stage using low-resolution dataset. However, both of them use multi-stage architecture, which has led to slow inference, making it difficult to meet some real-world application scenarios. And a more serious problem is the semantic incoherence between networks. Thus we aim to design a one-stage deep network to get rid of the above defects.

\section{UHR Saliency Detection Dataset}
\begin{figure}[t]

\centering
\includegraphics[width=1\columnwidth]{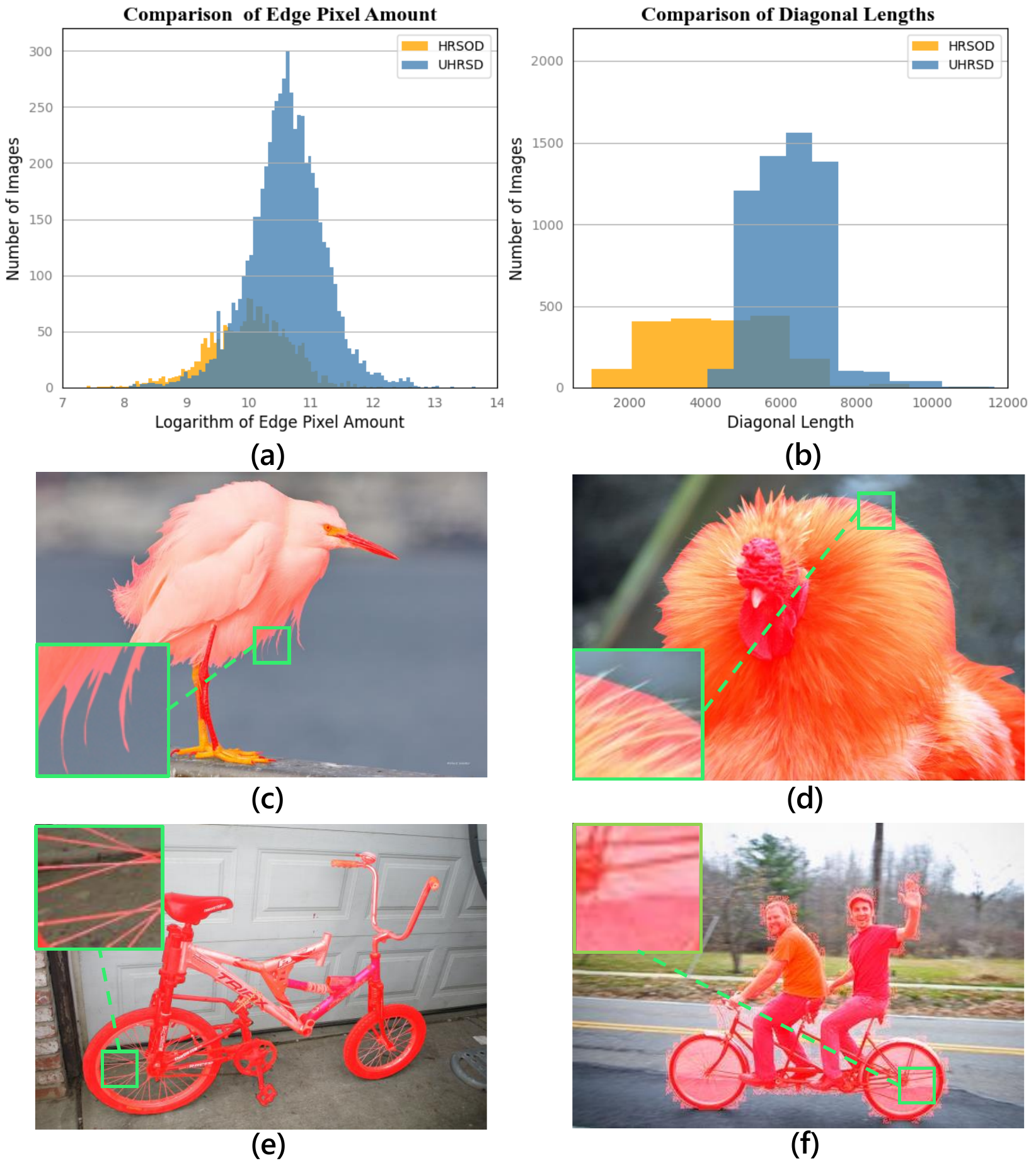}
\caption{Comparison of the results of the different methods. (a) Comparison of the logarithm of edge pixel amount between our UHRSD and HRSOD. (b) Comparison of the diagonal length between our UHRSD and HRSOD\cite{zeng2019towards} (c) Sample from our UHRSD. (d) Sample from HRSOD. (e) Sample from our UHRSD. (f) Sample from DUTS-TE. Best viewed by zooming in. }

\label{fig:dataset}
\end{figure}

\textbf{Available SOD datasets.} The existing common SOD datasets usually are in low-resolution (below $ 500 \times 500$ ). What's more, they have the following drawbacks for training high-resolution networks and evaluating high-quality segmentation results. Firstly the low resolution of the images results in insufficient detail information. Secondly, the quality of the edges of annotations is poor\cite{zeng2019towards} .Lastly, the finer level of annotations is dissatisfied, especially for hardcase annotations which are handled perfunctorily as shown in \cref{fig:dataset} (f).  The only available high-resolution dataset known is HRSOD\cite{zeng2019towards}. However, the number of high-resolution images
 in HRSOD is limited.

\textbf{UHRSD dataset.} For supervised learning, training data is obviously important. Before this, the only available high-resolution training set was only 1,610 images, and we experimentally discovered that training only on it was easy to overfit its data distribution, which significantly impacted the model's generalization ability. If the low-resolution datasets are mixed together for training, a lot of noise will be introduced to affect the performance of the high-resolution model. To relief the lack of high-resolution datasets for SOD, we contribute the Ultra High-Resolution for Saliency Detection (UHRSD) dataset with a total of 5,920 images in 4K($3840 \times 2160 $) or higher resolution, including 4,932 images for training and 988 images for testing. A total of 5,920 images were manually selected from websites (\eg Flickr Pixabay) with free copyright. Our dataset is diverse in terms of image scenes, with a balance of complex and simple salient objects of various size. Multiple participants in the constructing process to ensure accuracy of salient annotations. \cref{fig:dataset} illustrates the superiority of our UHRSD.  As shown in histogram \cref{fig:dataset} (a) (b), UHRSD datset is much larger than HRSOD datset and to the best of our knowledge is the largest dataset available. The large scale considerably alleviates the issues mentioned above when training high-resolution deep neural networks. In addition, the histogram \cref{fig:dataset}(b) shows that the size of images in UHRSD far exceeds that of the existing high-resolution dataset. Not only that, \cref{fig:dataset}(a) shows the number of pixels at the edges of our images also far surpasses the existing high-resolution dataset by a large margin, which means that UHRSD has richer and more challenging edge details. Lastly, through the comparison among \cref{fig:dataset}(c)-(f), it's evident that UHRSD also has a finer level of annotation for the hard cases than both existing high-resolution dataset and low-resolution dataset.

\label{sec:dataset}

\begin{figure*}[h]
  \centering
  \includegraphics[width=\linewidth]{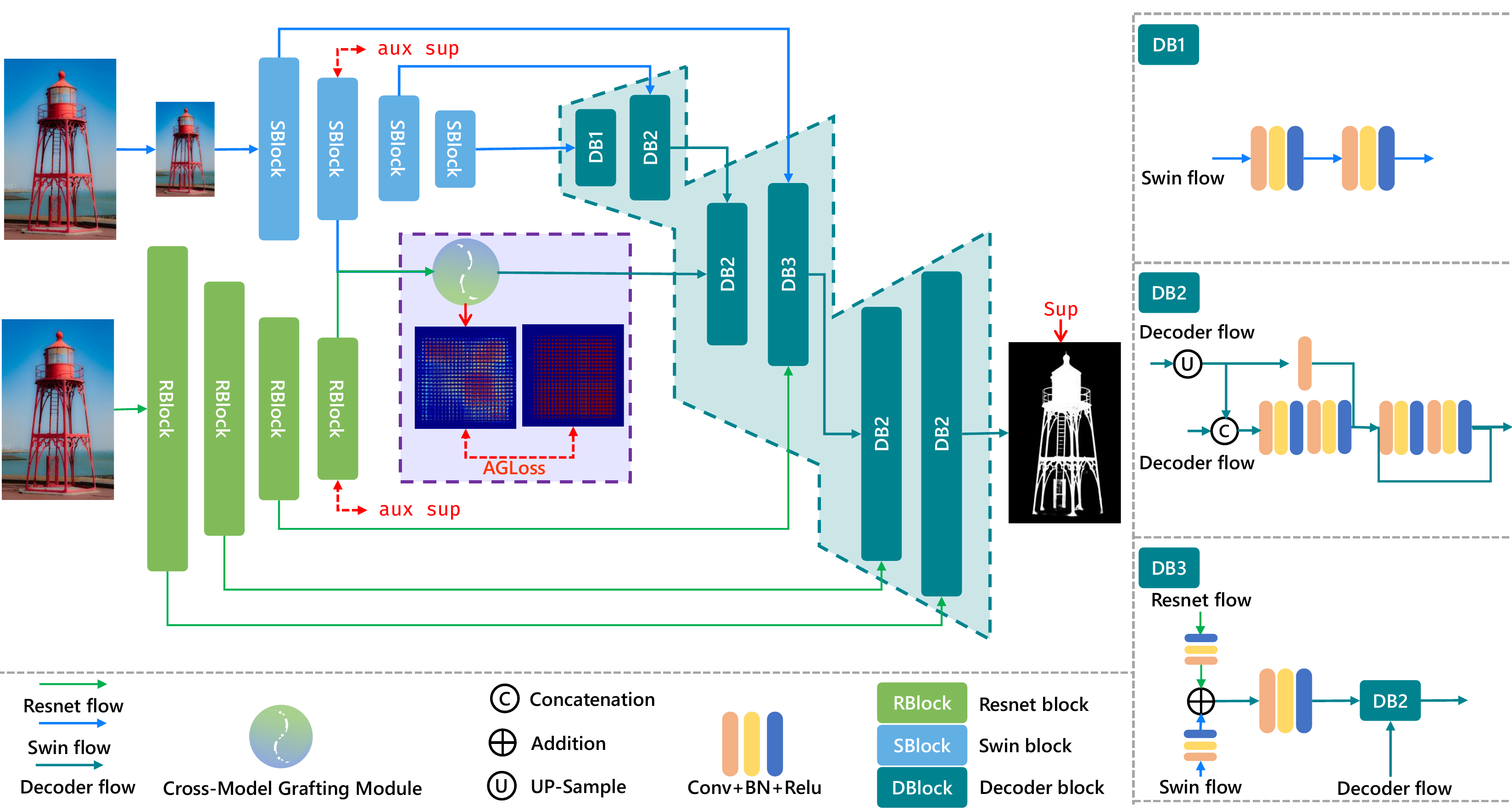}
  \caption{An overview of proposed Pyramid Grafting Network. Dual branches use Resnet and Swin transformer as encoder respectively. The DB$n$ is the Decoder block with $n$ input features, and the specific structure are shown on the right side. The two auxiliary supervisions are used to supervise the $RP$ and $SP$ mentioned in \cref{subsec:agl}.}
  \label{fig:frame}
\end{figure*}
\section{Methodology}

\subsection{Staggered Grafting Framework}
The architecture of proposed network is shown in \cref{fig:frame}. As can be seen, the network consists of two encoders and a decoder. To better perform the respective tasks of the two encoders, Swin transformer and Resnet-18 are chosen as encoders. This choice of combination was made for the consideration of balancing efficiency and effectiveness. On the one hand, the transformer encoder is able to get accurate global semantic information in the low-resolution case, and the convolutional encoder can get rich detail with the high-resolution input. On the other hand, variability in the features extracted by different models may be complementary to identify saliency more accurately.

During the encoding process, two encoders are fed with images of different resolutions in order to capture global semantic information and detailed information respectively in parallel. The decoding phase can be divided into three sub-stages, first Swin decoding, followed by grafted feature decoding and finally Resnet decoding in a staggered structure. The feature decoded in the second sub-stage is produced from Cross-Model Grafting Module (CMGM), where the global semantic information is grafted from Swin branch to Resnet branch. Also the CMGM process a matrix named $\mathrm{CAM}$ to be supervised. Reviewing the whole process, we construct a higher feature pyramid through two lower pyramid using  staggered connection structure as shown in \cref{fig:first}. In other word, the network achieves deeper sampling depth at low computational cost to adapt to the challenge caused by high-resolution input.

\subsection{Feature Extractors}
Countering the massive computational consumption and memory usage generated by high-resolution input, we choose Resnet-18 \cite{he2016deep} and Swin-B \cite{liu2021swin} as our backbones to balance performance and efficiency. For Resnet-18 encoder, five feature maps will be generated, which we denote the set as $\mathbb{R}$. The feature map extracted by top $7 \times 7 $ layer offers limited performance gains but consume huge computational effort, especially for high-resolution input. Thus the utilized features in $\mathbb{R}$ can be denoted as $\left\{ \bm{R}_i|i=2,3,4,5 \right\}$. Due to the down-sampling in every stage,  for input size $ H \times W$, the size of feature $\bm{R}_i$ is $ \frac{H}{2^i} \times \frac{W}{2^i} \times ({C}\times{2^{i-1}})$, where $({C}\times{2^i})$ is the channel of features. We remove the last stage while adopt the patch embedding feature of Swin transformer, which generates 4 features denoted as $\left\{ \bm{S}_i|i=1,2,3,4 \right\}$. Due to the nature that the embedding dim is fixed in transformer, the input size is $224 \times 224$ and the feature size in $\mathbb{S}$ is$\left\{ \frac{56}{2^{i-1}} \times \frac{56}{2^{i-1}} \times ({64}\times{2^i}) \right\}$ for $i=1,2,3$ and $14 \times 14 \times 512$ for $\bm{S_4}$.  The spatial size of $\bm{R}_5$ is close to $\bm{S}_2$, hence we chose to graft the features here.
\label{subsec:feature extracter}

\subsection{Cross-Model Grafting Module}
We propose Cross-Model Grafting Module(CMGM) to graft the feature $f_{{R}_5}$ and $f_{{S}_2}$ extracted by two different encoders. For feature $f_{{S}_2}$, due to the transformer's ability to capture information over long distances, it has global semantic information that is important for saliency detection. In contrast, CNNs perform well at extracting local information thus $f_{{R}_5}$ have relatively rich details. However, due to the contradiction between feature size and receptive field, in $f_{{R}_5}$ there will be many noises in the background.
\begin{figure}[t]
\centering
\includegraphics[width=1.0\columnwidth]{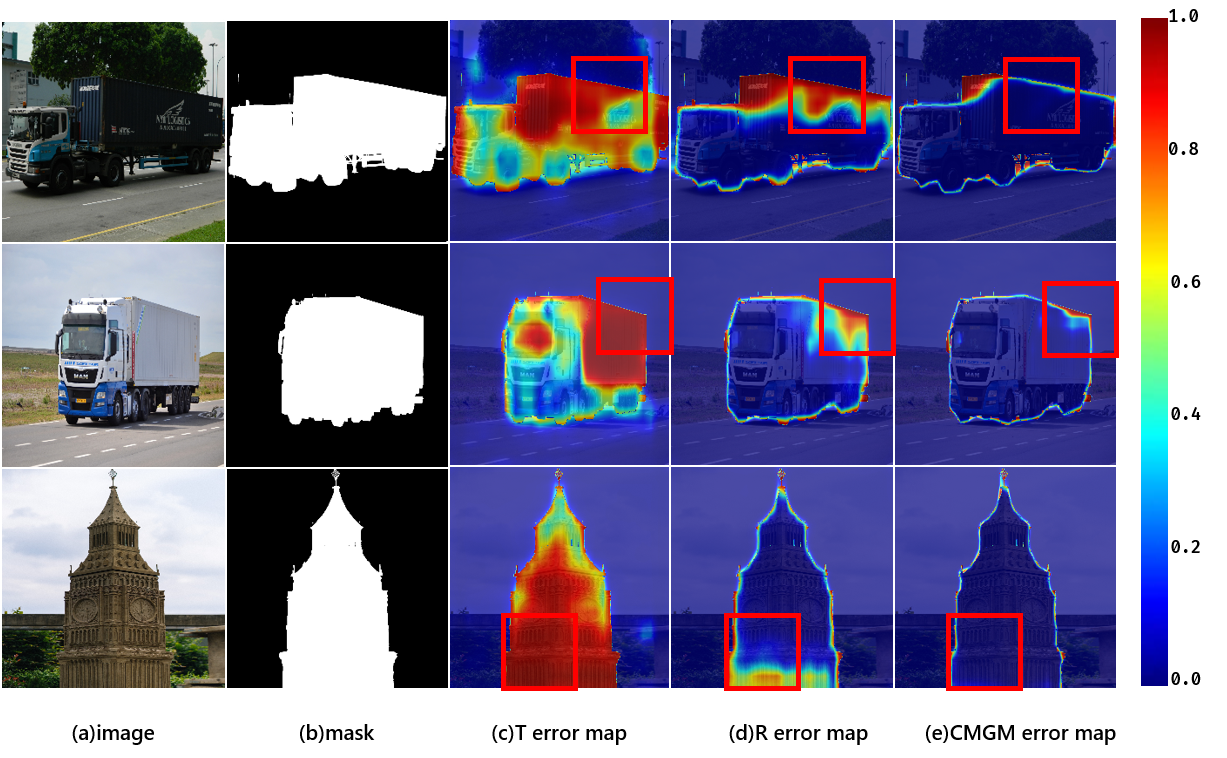}
\caption{Examples of error elimination through CMGM. (a)(b) are the images and ground truth masks. (c) and (d) are thee salient error map generated from Transformer branch and Resnet branch respectively. (e) shows the error map generated from CMGM.}
\label{fig:cmgm:errormap}
\end{figure}
For a salient prediction of a certain region, the predictions generated from different features can be
roughly summarized as three cases: (a) Both right, (b) Some of them right and (c) Both wrong. Existing fusion method using element-wise operation such as addition and multiplication may work for the first two cases. However, the element-wise operation and the convolutional operation focus on only limited local information, resulting fusion methods hardly remedy for common errors. Compared with the feature fusion, 
CMGM re-calculates the point-wise relationship between Resnet feature and Transformer feature, transferring the global semantic information from Transformer branch to Resnet branch so as to remedy the common errors. We calculate the error map by ${E}=|G-P| \in [0,1]$,where $G$ is the ground truth and $P$ is the salient prediction map generated by different branchs or CMGM. As shown in \cref{fig:cmgm:errormap}, the CMGM remedy the common error as expected.

\begin{figure}[t]

\centering
\includegraphics[width=0.9\columnwidth]{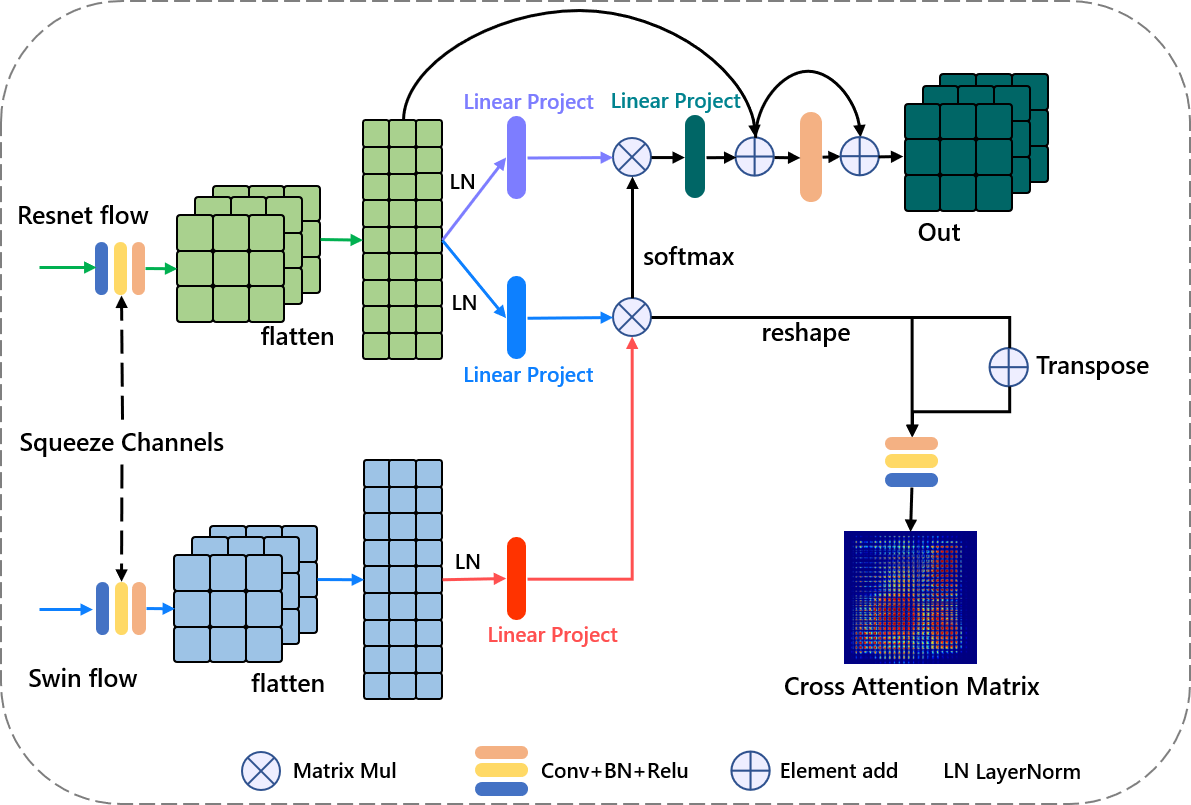}
\caption{Architecture of Cross-Model Grafting Module. }
\label{fig:cmgm}
\end{figure}

Specifically, in CMGM it first flatten the $f_{{R}_5} \in \upsilon^{H\times W\times C}$ to $f_{{R}}' \in \upsilon^{1\times C\times HW}$, and do the same to $f_{{{S}}_2}$ to get $f_{{S}}'$. Inspired by the multi-head self-attention mechanism, we apply layer normalization and linear projection on them respectively to get $f_{R}^q, f_{R}^v$ and $f_{S}^k$. We obtain $\bm{Z}$ by matrix multiplication, the process can be expressed as follows:
\begin{equation}
    \bm{Y} = \text {softmax}(f_{R}^q \times {f_{S}^k}^\mathrm{T})  ,
\end{equation}
\begin{equation}
    \bm{Z} = \bm{Y} \times f_{R}^v   ,
\end{equation}
then we input $\bm{Z}$ to the linear projection layer and reshape it back to size of $\upsilon^{H\times W\times C}$ before feeding into convolutional layer. Two shortcut connections were performed in the process as shown in \cref{fig:cmgm}. In addition, during the cross attention process, we generate Cross Attention Matrix based on $\bm{Y}$, which can be expressed as :

\begin{equation}
    \mathrm{CAM} = \text{ReLU}(\text{BN}(\text{Conv}(\bm{Y}+\bm{Y}^\mathrm{T})))   ,
\end{equation}
The detailed usage of $\mathrm{CAM}$ can be found in \cref{subsec:agl}.

\label{subsec:CMGM}

\subsection{Attention Guided Loss}
In order for CMGM  to better serve the purpose of transferring information from the Transformer branch to the Renset branch, we design the Attention Guided Loss to supervise the Cross Attention Matrix explicitly. We argue that the Cross Attention Matrix should be similar to the attention matrix generated from ground truth, because the salient features should have a higher similarity, in other words the dot product should has a larger activation value.
As shown in \cref{fig:agl1} given a salient map $\bm{M}$ with size $H \times W$, we first flatten it to $\bm{M}'$ with size $1 \times HW$. Then we apply matrix multiplication on $\bm{M}'$ to obtain corresponding attention matrix $\bm{M}^{a}$. The process can be denoted as $\bm{M}^{a} = \mathcal {F} \left( \bm{M} \right)$ and the value of $\bm{M}^{a}_{xy}$ can be expressed as 

\begin{equation}
   \bm{M}^{a}_{xy} = {{\bm{M}'}^\mathrm{T}}_{x} \times \bm{M}'_{y} ,
    \label{eq:attention matrix}
\end{equation}

Then we use the transformation $\mathcal {F}(\cdot)$ to construct $G^{a},RP^{a},SP^{a}$, where $G$ is the ground truth map, $RP$ and $SP$ are salient prediction map generated from feature $\bm{R}_5$ and $\bm{S}_2$ respectively. We propose the Attention Guided Loss based on weighted binary cross entropy (wBCE) to supervise the Cross Attention Matrix $\mathrm{CAM}$ generated from CMGM shown in \cref{fig:cmgm}. The BCE \cite{de2005tutorial} can be written as:

\begin{figure}[ht]
\centering
\includegraphics[width=0.9\columnwidth]{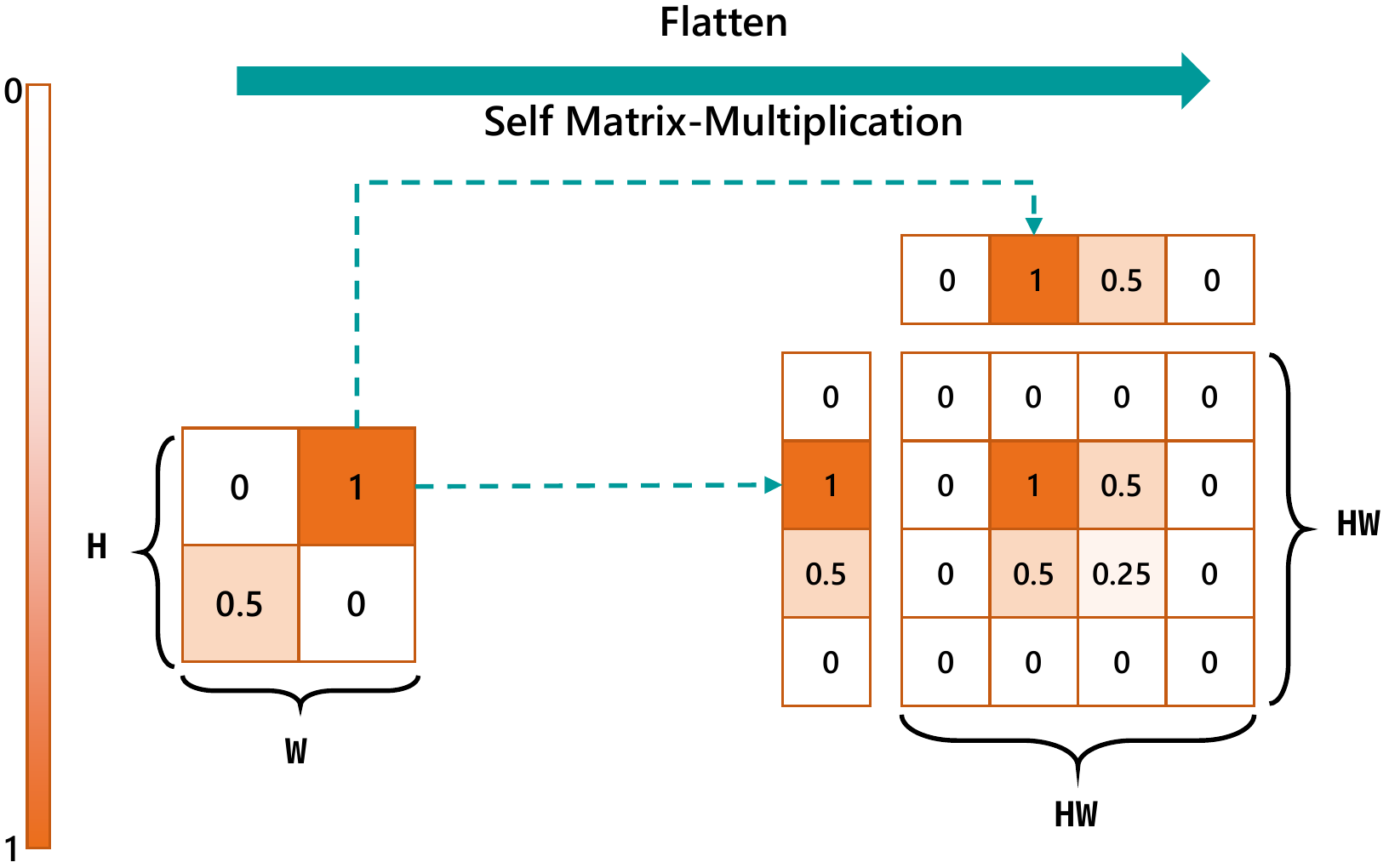}
\caption{The construction of attention matrix. The operation is used to create the target and weights for proposed AGL. }
\label{fig:agl1}
\end{figure}

\begin{equation}
    \ell_{bce}(G_{xy},P_{xy}) = 
    \begin{cases}
    \log(P_{xy})& G_{xy}=1 \\
    \log(1-P_{xy})& G_{xy}=0
    \end{cases} ,
    \label{eq:bce}
\end{equation}
where $G_{xy}$ is the ground truth label of the pixel $(x,y)$, and $P_{xy}$ is the predicted probability in predicted map and both of them are in range$[0,1]$. Then our $\mathcal{L}_{AG}$ can be expressed as:
\begin{equation}
    \mathcal{L}_{AG}=-\frac{\sum\limits^H_{i=1}\sum\limits^W_{j=1}(1+\beta\omega_{ij})\cdot \ell_{bce}(G^a_{ij},{\mathrm{CAM}}_{ij})}{\sum\limits^H_{i=1}\sum\limits^W_{j=1}(1+\beta\omega_{ij})} ,
    \label{eq:agl}
\end{equation}
where $\beta$ is a hyperparameter to adjust impact of the weight $\omega$ \cref{eq:omega}. In the \cref{eq:agl}, the $\ell_{bce}$ on each pixel is assigned with a weight $\beta\omega_{ij}$. The use of weight $\omega$ serves two purposes: (1) The degree of positive and negative sample imbalance is squared due to the matrix multiplication.(2) As described in \cref{subsec:CMGM}, we want to remedy the errors common to both of branches. When $\beta \omega$ equals  $0$, the \cref{eq:agl} becomes usual binary cross entropy loss $\mathcal{L}_{bce}$. The weight $\omega$ can be calculated by:

\begin{equation}\small
    \omega_{ij} = \frac{1}{2}(\mid(G^a_{ij}-RP^a_{ij})\mid +\mid(G^a_{ij}-SP^a_{ij})\mid)+1    ,
    \label{eq:omega}
\end{equation}
where $RP^a$ and $SP^a$ are the attention matrix of $RP$ and $SP$ defined above.

What's more, we also apply the widely-used IoU loss \cite{mattyus2017deeproadmapper} to pay more attention to the global structure of the image as suggested by \cite{qin2019basnet}. The IoU loss $\mathcal{L}_{iou}$ can be referred to supplementary materials.
In the end, our total loss can be expressed as follow:
\begin{equation}
    \mathcal{L}_{total} = \mathcal{L}_{b+ i}^{P}+\mathcal{L}_{AG}+\frac{1}{8}(\mathcal{L}_{b+ i}^{auxiliary}),
    \label{eq:total}
\end{equation}
where $\mathcal{L}_{b+i}=\mathcal{L}_{bce}+\mathcal{L}_{iou}$, and $\mathcal{L}^{auxiliary}_{b+i}$ is $\mathcal{L}_{b+i}$ applied on the $RP$ and $SP$.

\label{subsec:agl}
\section{Experiments}

\subsection{Datasets and Evaluation Metrics}

\textbf{High-Resolution Datasets.} 
The high-resolution datasets available are UHRSD (4,932 images for training and 988 for testing) , HRSOD\cite{zeng2019towards} (1,610 images for training and 400 for testing). Followed by \cite{zeng2019towards,tang2021disentangled}, we also use the DAVIS-S for evaluation. 

\textbf{Low-Resolution Datasets.} 
DUTS-TR \cite{wang2017learning} is used to train the model. In addition, we also evaluate our method on widely-used benchmark datasets: ECSSD \cite{yan2013hierarchical} with 1,000 images, DUT-OMRON \cite{yang2013saliency} with 5,168 images, PASCAL-S\cite{li2014secrets} with 850 images, DUTS-TE \cite{wang2017learning} with 5,019 images and HKU-IS\cite{li2015visual} with 4,447 images.

\textbf{Evaluation Metrics.}
We use following metrics to evaluate the performance of all methods. Firstly, Mean Absolute Error (MAE), defined as \cref{eq:mae} where $P$ is the prediction map and $G$ is the ground truth.  
The second is Max $\small F\text{-measure}$ ($F_{\beta}^{Max}$), which can be calculated by
$F_\beta = \frac{(1+\beta^2) \cdot \text{precision} \cdot \text{recall}}{\beta^2 \cdot \text{precision} \cdot \text{recall}}$, where $\beta^2$ is set to 0.3 as suggested in \cite{borji2015salient}. 
Then we adopt Structural similarity Measure ($S_m$) \cite{fan2017structure} and $E\text{-measure}$($E_\xi$) \cite{fan2018enhanced} as many other methods\cite{wei2020f3net,ma2021pyramidal}. 
At last, to better evaluate the boundary quality which is important in High-resolution Saliency Detection\cite{zeng2019towards,tang2021disentangled}, we adopt the Boundary Displacement Error (BDE) to evaluate the result of high-resolution datasets, where lower values means better boundary quality.
\begin{equation} \small
    \mathrm{MAE}=\frac{1}{H \times W} \sum\limits_{i=1}^{H}\sum\limits_{j=1}^W|P_{ij}-G_{ij}| .
    \label{eq:mae}
\end{equation}

\subsection{Implementation Details}

\begin{table*}[t]
\caption{Quantitative comparisons with state-of-the-art SOD models on five benchmark datasets in terms of  max F-measure, MAE , E-measure, S-measure and BDE. The best two results are shown in red and green, respectively. D: trained on DUTS-TR, HD: trained on DUTS-TR and HRSOD-TR, UH: trained on UHRSD-TR and HRSOD-TR . The best two results are in {\color[HTML]{FE0000} red} and {\color[HTML]{34FF34} green} fonts.}

\label{table:performance}
\LARGE
\renewcommand\arraystretch{1.3}
\centering
\setlength\tabcolsep{6pt}
\resizebox{0.95\textwidth}{!}{%

\begin{tabular}{lccccccccccccccccccccccc}
\toprule[2pt]
\multicolumn{1}{l|}{}                                  & \multicolumn{5}{c|}{\textbf{HRSOD-TE}}                                                                                                                                         & \multicolumn{5}{c|}{\textbf{DAVIS-S}}                                                                                                                                          & \multicolumn{5}{c|}{\textbf{UHRSD-TE}}                                                                                                                                         & \multicolumn{4}{c|}{\textbf{DUT-OMRON}}                                                                                                        & \multicolumn{4}{c}{\textbf{DUTS-TE}}                                                                                      \\ \cline{2-24} 
\multicolumn{1}{l|}{\multirow{-2}{*}{\textbf{Method}}} & $F_\beta^{Max}$                         & $\mathrm{MAE}$                          & $E_\xi$                           & $S_m$                           & \multicolumn{1}{c|}{$\mathrm{BDE}$}                           & $F_\beta^{Max}$                         & $\mathrm{MAE}$                          & $E_\xi$                           & $S_m$                           & \multicolumn{1}{c|}{$\mathrm{BDE}$}                           & $F_\beta^{Max}$                         & $\mathrm{MAE}$                          & $E_\xi$                           & $S_m$                           & \multicolumn{1}{c|}{$\mathrm{BDE}$}                           & $F_\beta^{Max}$                         & $\mathrm{MAE}$                          & $E_\xi$                           & \multicolumn{1}{c|}{$S_m$}                           & $F_\beta^{Max}$                         & $\mathrm{MAE}$                          & $E_\xi$                           & $S_m$                           \\ \toprule[2pt]
\multicolumn{1}{l|}{\textbf{CPD$_{19}$}}                      & .867                        & .041                        & .891                        & .881                        & \multicolumn{1}{c|}{62.066}                        & .871                        & .029                        & .921                        & .893                        & \multicolumn{1}{c|}{33.971}                        & .894                        & .055                        & .884                        & .878                        & \multicolumn{1}{c|}{32.587}                        & .797                        & .056                        & .866                        & \multicolumn{1}{c|}{.825}                        & .865                        & .043                        & .887                        & .869                        \\
\multicolumn{1}{l|}{\textbf{SCRN$_{19}$}}                     & .880                        & .042                        & .887                        & .888                        & \multicolumn{1}{c|}{75.696}                        & .893                        & .027                        & .911                        & .902                        & \multicolumn{1}{c|}{46.592}                        & .904                        & .051                        & .880                        & .887                        & \multicolumn{1}{c|}{40.176}                        & .811                        & .056                        & .863                        & \multicolumn{1}{c|}{.837}                        & .888                        & .040                        & .888                        & .885                        \\
\multicolumn{1}{l|}{\textbf{DASNet$_{20}$}}                   & .893                        & .032                        & .925                        & .897                        & \multicolumn{1}{c|}{69.310}                        & .902                        & .020                        & .949                        & .911                        & \multicolumn{1}{c|}{26.761}                        & .914                        & .045                        & .892                        & .889                        & \multicolumn{1}{c|}{35.044}                        & {\color[HTML]{34FF34} .827} & .050                        & .877                        & \multicolumn{1}{c|}{.845}                        & .895                        & .034                        & .908                        & .894                        \\
\multicolumn{1}{l|}{\textbf{F3Net$_{20}$}}                    & .900                        & .035                        & .913                        & .897                        & \multicolumn{1}{c|}{65.757}                        & .915                        & .020                        & .940                        & .914                        & \multicolumn{1}{c|}{44.760}                        & .909                        & .046                        & .887                        & .890                        & \multicolumn{1}{c|}{39.612}                        & .813                        & .053                        & .871                        & \multicolumn{1}{c|}{.838}                        & .891                        & .035                        & .902                        & .888                        \\
\multicolumn{1}{l|}{\textbf{GCPA$_{20}$}}                     & .889                        & .036                        & .898                        & .898                        & \multicolumn{1}{c|}{74.900}                        & .922                        & .020                        & .934                        & .929                        & \multicolumn{1}{c|}{39.160}                        & .912                        & .047                        & .886                        & .896                        & \multicolumn{1}{c|}{35.947}                        & .812                        & .056                        & .860                        & \multicolumn{1}{c|}{.839}                        & .888                        & .038                        & .891                        & .891                        \\
\multicolumn{1}{l|}{\textbf{ITSD$_{20}$}}                     & .896                        & .036                        & .912                        & .898                        & \multicolumn{1}{c|}{87.946}                        & .899                        & .022                        & .922                        & .909                        & \multicolumn{1}{c|}{68.256}                        & .911                        & .045                        & .895                        & .897                        & \multicolumn{1}{c|}{41.174}                        & .821                        & .061                        & .863                        & \multicolumn{1}{c|}{.840}                        & .883                        & .041                        & .895                        & .885                        \\
\multicolumn{1}{l|}{\textbf{LDF$_{20}$}}                      & .904                        & .032                        & .919                        & .904                        & \multicolumn{1}{c|}{58.714}                        & .911                        & .019                        & .947                        & .922                        & \multicolumn{1}{c|}{35.447}                        & .913                        & .047                        & .891                        & .888                        & \multicolumn{1}{c|}{33.775}                        & .820                        & .051                        & .873                        & \multicolumn{1}{c|}{.838}                        & .898                        & .034                        & .910                        & .892                        \\
\multicolumn{1}{l|}{\textbf{CTD$_{21}$}}                      & .905                        & .032                        & .921                        & .905                        & \multicolumn{1}{c|}{63.907}                        & .904                        & .019                        & .938                        & .911                        & \multicolumn{1}{c|}{42.832}                        & .917                        & .043                        & .898                        & .897                        & \multicolumn{1}{c|}{33.835}                        & .826                        & .052                        & .875                        & \multicolumn{1}{c|}{.844}                        & .897                        & .034                        & .909                        & .893                        \\
\multicolumn{1}{l|}{\textbf{PFS$_{21}$}}                      & .911                        & .033                        & .922                        & .906                        & \multicolumn{1}{c|}{63.537}                        & .916                        & .019                        & .946                        & .923                        & \multicolumn{1}{c|}{30.612}                        & .918                        & .043                        & .896                        & .897                        & \multicolumn{1}{c|}{37.387}                        & .823                        & .055                        & .875                        & \multicolumn{1}{c|}{.842}                        & .896                        & .036                        & .902                        & .892                        \\
\multicolumn{1}{l|}{\textbf{HRSOD-DH$_{19}$}}                 & .905                        & .030                        & .934                        & .896                        & \multicolumn{1}{c|}{88.017}                        & .899                        & .026                        & .955                        & .876                        & \multicolumn{1}{c|}{44.359}                        & -                            & -                            & -                            & -                            & \multicolumn{1}{c|}{-}                             & .743                        & .065                        & .831                        & \multicolumn{1}{c|}{.762}                        & .835                        & .050                        & .885                        & .824                        \\
\multicolumn{1}{l|}{\textbf{DHQSOD-DH$_{21}$}}                & .922                        & .022                        & {\color[HTML]{FE0000} .947} & .920                        & \multicolumn{1}{c|}{{\color[HTML]{34FF34} 46.495}} & .938                        & {\color[HTML]{34FF34} .012} & .947                        & .920                        & \multicolumn{1}{c|}{{\color[HTML]{34FF34} 14.266}} & -                            & -                            & -                            & -                            & \multicolumn{1}{c|}{-}                             & .820                        & {\color[HTML]{FE0000} .045} & .873                        & \multicolumn{1}{c|}{.836}                        & .900                        & .031                        & .919                        & .894                        \\ \toprule \hline
\multicolumn{24}{c}{\textbf{Our PGNet}}                                                                                                                                                                                                                                                                                                                                                                                                                                                                                                                                                                                                                                                                                                                                                                                                                                               \\ \midrule \hline
\multicolumn{1}{l|}{\textbf{Ours-D}}                   & .931                        & {\color[HTML]{34FF34} .021} & .944                        & .930                        & \multicolumn{1}{c|}{46.923}                        & .936                        & .015                        & .947                        & .935                        & \multicolumn{1}{c|}{34.957}                        & .931                        & .037                        & .904                        & .912                        & \multicolumn{1}{c|}{32.300}                        & {\color[HTML]{FE0000} .835} & {\color[HTML]{FE0000} .045} & {\color[HTML]{FE0000} .887} & \multicolumn{1}{c|}{{\color[HTML]{34FF34} .855}} & {\color[HTML]{34FF34} .917} & {\color[HTML]{FE0000} .027} & {\color[HTML]{34FF34} .922} & {\color[HTML]{34FF34} .911} \\
\multicolumn{1}{l|}{\textbf{Ours-DH}}                  & {\color[HTML]{34FF34} .937} & {\color[HTML]{FE0000} .020} & {\color[HTML]{34FF34} .946} & {\color[HTML]{34FF34} .935} & \multicolumn{1}{c|}{{\color[HTML]{FE0000} 45.292}} & {\color[HTML]{34FF34} .950} & {\color[HTML]{34FF34} .012} & {\color[HTML]{34FF34} .975} & {\color[HTML]{34FF34} .948} & \multicolumn{1}{c|}{14.463}                        & {\color[HTML]{34FF34} .935} & {\color[HTML]{34FF34} .036} & {\color[HTML]{34FF34} .905} & {\color[HTML]{34FF34} .912} & \multicolumn{1}{c|}{{\color[HTML]{34FF34} 32.008}} & {\color[HTML]{FE0000} .835} & {\color[HTML]{34FF34} .046} & {\color[HTML]{FE0000} .887} & \multicolumn{1}{c|}{{\color[HTML]{FE0000} .858}} & {\color[HTML]{FE0000} .919} & {\color[HTML]{34FF34} .028} & {\color[HTML]{FE0000} .925} & {\color[HTML]{FE0000} .912} \\
\multicolumn{1}{l|}{\textbf{Ours-UH}}                  & {\color[HTML]{FE0000} .945} & {\color[HTML]{FE0000} .020} & {\color[HTML]{34FF34} .946} & {\color[HTML]{FE0000} .938} & \multicolumn{1}{c|}{57.147}                        & {\color[HTML]{FE0000} .957} & {\color[HTML]{FE0000} .010} & {\color[HTML]{FE0000} .979} & {\color[HTML]{FE0000} .954} & \multicolumn{1}{c|}{{\color[HTML]{FE0000} 12.725}} & {\color[HTML]{FE0000} .949} & {\color[HTML]{FE0000} .026} & {\color[HTML]{FE0000} .916} & {\color[HTML]{FE0000} .935} & \multicolumn{1}{c|}{{\color[HTML]{FE0000} 30.019}} & .772                        & .058                        & {\color[HTML]{34FF34} .884} & \multicolumn{1}{c|}{.786}                        & .871                        & .038                        & .897                        & .859                        \\ \bottomrule[2pt]
\end{tabular}%

}
\end{table*}

\begin{figure*}[t]
  \centering
  \includegraphics[width=1.0\linewidth]{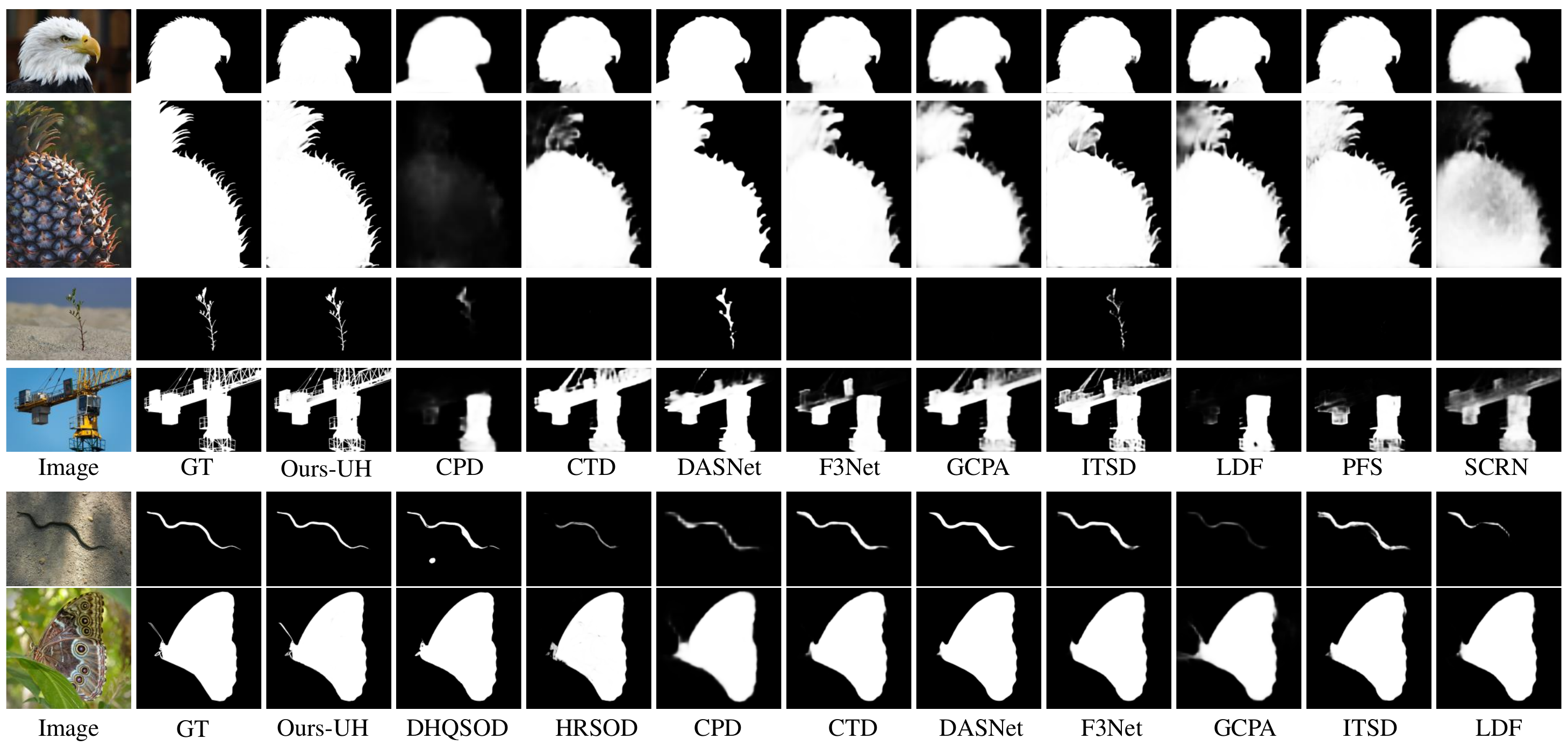}
  \caption{Visual comparison between our method and SOTA methods. The first four lines are from our UHRSD-TE and the next two lines are from HRSOD-TE. Best viewed by zooming in.}
  \label{fig:visual comparison}
\end{figure*}

We use Pytorch\cite{paszke2017automatic} to implement our model and two RTX 2080Ti GPUs are used for accelerating training. We choose Resnet-18 \cite{he2016deep} and Swin-B\_224 \cite{liu2021swin} as the backbone for convolutional branch and transformer branch respectively. 
The whole network is trained end-to-end by using stochastic gradient descent (SGD). We set the maximum learning rate to 0.003 for Swin backbone and 0.03 for others. The learning rate first increases then decays during the training process, what's more Momentum and weight decay are set to 0.9 and 0.0005 respectively. Batchsize is set to 16 and maximum epoch is set to 32. For data augmentation, we use random flip, crop and multi-scale input images \cite{tang2021disentangled,qin2019basnet,zhao2019egnet}. In order to make fair comparisons and fully demonstrate the attributes of our UHRSD, we take three combinations of available datasets to train our model: (1) DUTS-TR (2) DUTS-TR+HRSOD-TR (3) UHRSD-TR+HRSOD-TR.
During testing, each image is resized to $1024 \times 1024$ and then fed into the network without any post-processing(\eg CRF\cite{krahenbuhl2011efficient}).

\begin{table}[]
\caption{Comparison of different architectures and compositions.}

\renewcommand\arraystretch{1}
\label{table:ablation1}
\centering
\resizebox{\linewidth}{!}{%
\begin{tabular}{l|cccc}
\toprule[2pt]
\multicolumn{1}{c|}{\multirow{2}{*}{Composition}} & \multicolumn{4}{c}{HRSOD-TE} \\ \cline{2-5} 
\multicolumn{1}{c|}{}                            & $F_\beta^{Max}$                         & $\mathrm{MAE}$                          & $E_\xi$                           & $S_m$   \\ \toprule
baseline\_Resnet-18                               & .878  & .051  & .875  & .871 \\
baseline\_Swin                                   & .915  & .027  & .937  & .921 \\
baseline\_R+S+CMGM                               & .940  & .023  & .944  & .936 \\
baseline\_R+S+CMGM+AGL                           & .945  & .020  & .946  & .938 \\ \bottomrule[2pt]
\end{tabular}%
}
\end{table}

\subsection{Comparison with the State-of-the-arts}
We compare our proposed PGNet with 11 SOTA methods, including CPD \cite{wu2019cascaded}, SCRN \cite{wu2019stacked}, DASNet \cite{zhao2020depth}, F3Net \cite{wei2020f3net}, GCPA \cite{chen2020global}, ITSD \cite{zhou2020interactive}, LDF \cite{wei2020label}, CTD \cite{zhao2021complementary}, PFS \cite{ma2021pyramidal}, HRSOD \cite{zeng2019towards}, DHQSOD \cite{tang2021disentangled}, where HRSOD and DHQSOD are designed for high-resolution salient object detection.  All of the above methods use Resnet-50\cite{he2016deep} as the backbone except for HRSOD which uses VGG16 \cite{simonyan2014very}. And all of them are trained on DUTS-TR \cite{wang2017learning} dataset, except for the marked ones like HRSOD-DH and DHQSOD-DH, which are trained on the mixed dataset (HRSOD \cite{zeng2019towards} and DUTS-TR). For a fair comparison, we use either the available implementations or the saliency maps provided by the authors. It's worth noting that the vacant lines in \cref{table:performance} are caused by the fact that one of them is not available so far and the other not being consistent with our test environment.
\newpage
\textbf{Quantitative Comparison.}
As mentioned above, for fair comparison we use three settings of train set. As can be seen in \cref{table:performance}, the results of training on either only DUTS-TR or mix of DUTS-TR and HRSOD-TR exceed the SOTA by a large margin on both high-resolution and low-resolution test sets. When using the mixed dataset DUTS-HRSOD, our method has significantly improved on high-resolution datasets. There may be discrepancy between the distribution of high-resolution and low-resolution data. This is further supported by the results of training on the  UHRSD-HRSOD mixed dataset, where the performance of the high-resolution dataset is significantly improved, especially for UHRSD-TE. This demonstrates that the annotation bias of high-resolution datasets differing from low-resolution datasets has a promotional effect on supervised high-resolution saliency detection method, which is the reason why high-resolution training data with high-quality annotation is in great demand.

\textbf{Visual Comparison.}
To exhibit the characteristics of high-resolution dataset and the superiority of our method on it, \cref{fig:visual comparison} shows representative examples of visual comparison of ours with respect to others. As can be seen, our method can capture details well and produce clear boundary (row 1 and 2). More than the high quality boundary, another significant aspect of high-resolution SOD is the ability to segment objects with small sturctures that are easily to overlook in low-resolution cases (row 3, 5 and 6). This also demonstrates the superiority that our method makes the process one-stage. What's more, our method works well even in some extremly complex scenarios such as row 4.

\subsection{Ablation Study}
To better illustrate the nature of proposed method for high-resolution images, the ablation studies are based on the settings of  \textbf{Ours-UH}, which is trained on the mixed dataset UHRSD-TR and HRSOD-TR.

\textbf{Ablation Study for Compositions.} To prove the effectiveness of proposed feature grafting method including the CMGM and AGL, we report the quantitative performance in \cref{table:ablation1}. The baseline\_Resnet-18 and baseline\_Swin represent the widely-used U-shape network with Resnet-18 backbone and Swin backbone respectively. As can be seen in row 3, our proposed staggered architecture and Cross-Model Grafting Module inherits the strengths of both models. What's more, under the guiding role of AGL, performance has been further improved.


\begin{table}[]
\caption{Performance with the different grafted features. $\bm{R}_i$ denotes the $i$th feature of $\mathbb{R}$ defined in \cref{subsec:feature extracter}, and $\bm{S}_i$ is similar.}
\label{table:ablation2}
\centering
\renewcommand\arraystretch{1}
\resizebox{\linewidth}{!}{%
\begin{tabular}{l|ccc|ccc}
\toprule[2pt]
\multirow{2}{*}{Feature Pair} & \multicolumn{3}{c|}{HRSOD-TE} & \multicolumn{3}{c}{UHRSD-TE} \\ \cline{2-7} 
                           & $F_\beta^{Max}$                         & $\mathrm{MAE}$                          & $E_\xi$       & $F_\beta^{Max}$                         & $\mathrm{MAE}$                          & $E_\xi$      \\ \toprule
$R_5-S_4$                     & .913     & .029     & .922    & .935     & .031    & .907    \\
$R_5-S_3$                     & .939     & .022     & .937    & .947     & .026    & .912    \\
$R_5-S_2$            & .945     & .020     & .946    & .949     & .026    & .916    \\
$R_5-S_1$                     & .937     & .022     & .935    & .947     & .026    & .910    \\ \bottomrule[2pt]
\end{tabular}%
}
\end{table}

\textbf{Ablation Study for Grafting Position.}
To investigate the impact of grafting position on the network performance, we conduct a series of experiments with different grafting feature pairs. As shown in \cref{table:ablation2}, starting with the alignment of the last stage of two encoders, the performance gradually improves as the number of staggered layers increase until reaching the best at the pair $\bm{R}_5-\bm{S}_2$. This may be due to the spatial size of feature maps. When the sizes are close, the spatial information in the features extracted from two models corresponds to each other, which in turn promotes the feature grafting.

\section{Limitation}
Our method is simple and fast for one-stage high-resolution saliency detection, but the training process is still quite demanding on GPU memory usage, resulting in a high cost of training. What's more, though our method has already a superior input resolution compared to previous SOD methods, the input resolution is not unlimited. For excessive resolution such as 4K, images need to be down-sampled first before input.

\section{Conclusion}
In this paper, we propose the Pyramid Grafting Network for one-stage high-resolution salient object detection. The proposed staggered grafting patterns effectively exploit the advantages of each of the two existing different encoder. In addtion the proposed Cross-Model Grafting Modules and Attention Guided Loss cooperate with each other to inherit the advantages and remedy the common defects of the CNN and transformer. It is worth noting that we contribute the first 4K resolution SOD dataset for advancing future studies in high-resolution SOD. Extensive experiments demonstrates that our method not only outperforms the state-of-the-art methods but also is able to produce high-resolution saliency predictions fast and accurately.

\noindent\textbf{Acknowledgments:} This work was supported by the National Natural Science Foundation of China under Grant 62132002, Grant 61922006 and Grant 62102206.

{\small
\bibliographystyle{ieee_fullname}
\bibliography{PaperForReview}
}

\end{document}